# Neural Distributed Autoassociative Memories: A Survey

**V.I. Gritsenko**, **D.A. Rachkovskij**, **A.A. Frolov**, **R. Gayler, D. Kleyko, E. Osipov**

**Abstract**

**Introduction.** Neural network models of autoassociative, distributed memory allow storage and retrieval of many items (vectors) where the number of stored items can exceed the vector dimension (the number of neurons in the network). This opens the possibility of a sublinear time search (in the number of stored items) for approximate nearest neighbors among vectors of high dimension.

**The purpose of this paper** is to review models of autoassociative, distributed memory that can be naturally implemented by neural networks (mainly with local learning rules and iterative dynamics based on information locally available to neurons).

**Scope.** The survey is focused mainly on the networks of Hopfield, Willshaw and Potts, that have connections between pairs of neurons and operate on sparse binary vectors. We discuss not only autoassociative memory, but also the generalization properties of these networks. We also consider neural networks with higher-order connections and networks with a bipartite graph structure for non-binary data with linear constraints.

**Conclusions.** In conclusion we discuss the relations to similarity search, advantages and drawbacks of these techniques, and topics for further research. An interesting and still not completely resolved question is whether neural autoassociative memories can search for approximate nearest neighbors faster than other index structures for similarity search, in particular for the case of very high dimensional vectors.

# Информатика и информационные технологии

**V.I. GRITSENKO**[1], Corresponding Member of NAS of Ukraine, Director,
e-mail: vig@irtc.org.ua
**D.A. RACHKOVSKIJ**[1], Dr (Engineering), Leading Researcher,
Dept. of Neural Information Processing Technologies,
e-mail: dar@infrm.kiev.ua
**A.A. FROLOV**[2], Dr (Biology), Professor,
Faculty of Electrical Engineering and Computer Science FEI,
e-mail: docfact@gmail.com
**R. GAYLER**[3], PhD (Psychology), Independent Researcher,
e-mail: r.gayler@gmail.com
**D. KLEYKO**[4] graduate student,
Department of Computer Science, Electrical and Space Engineering,
e-mail: denis.kleyko@ltu.se
**E. OSIPOV**[4] PhD (Informatics), Professor,
Department of Computer Science, Electrical and Space Engineering,
e-mail: evgeny.osipov@ltu.se

[1] International Research and Training Center for Information Technologies and Systems of the NAS of Ukraine and of Ministry of Education and Science of Ukraine, ave. Acad. Glushkova, 40, Kiev, 03680, Ukraine
[2] Technical University of Ostrava, 17 listopadu 15, 708 33 Ostrava-Poruba, Czech Republic
[3] Melbourne, VIC, Australia
[4] Lulea University of Technology, 971 87 Lulea, Sweden

# NEURAL DISTRIBUTED AUTOASSOCIATIVE MEMORIES: A SURVEY

***Introduction.*** *Neural network models of autoassociative, distributed memory allow storage and retrieval of many items (vectors) where the number of stored items can exceed the vector dimension (the number of neurons in the network). This opens the possibility of a sublinear time search (in the number of stored items) for approximate nearest neighbors among vectors of high dimension.*

*__The purpose of this paper__ is to review models of autoassociative, distributed memory that can be naturally implemented by neural networks (mainly with local learning rules and iterative dynamics based on information locally available to neurons).*

*__Scope.__ The survey is focused mainly on the networks of Hopfield, Willshaw and Potts, that have connections between pairs of neurons and operate on sparse binary vectors. We*

*V.I. Gritsenko, D.A. Rachkovskij, A.A. Frolov, R. Gayler, D. Kleyko, E. Osipov*

*discuss not only autoassociative memory, but also the generalization properties of these networks. We also consider neural networks with higher-order connections and networks with a bipartite graph structure for non-binary data with linear constraints.*

***Conclusions.*** *In conclusion we discuss the relations to similarity search, advantages and drawbacks of these techniques, and topics for further research. An interesting and still not completely resolved question is whether neural autoassociative memories can search for approximate nearest neighbors faster than other index structures for similarity search, in particular for the case of very high dimensional vectors.*

## INTRODUCTION

In this paper, we review some artificial neural network variants of distributed autoassociative memories (denoted by Neural Associative Memory, NAM) [1–159].

Varieties of associative memory [93] (or content addressable memory) can be considered as index structures performing some types of similarity search. In autoassociative memory, the output is the word of memory, most similar to the key at the input. We restrict our initial attention to systems where the key and memory words are binary vectors. Therefore, autoassociative memory answers nearest neighbor queries for binary vectors.

In distributed memory, different vectors (items to be stored) are stored in shared memory cells. That is, each item to be stored consists of a pattern of activation across (potentially) all the memory cells of the system and each memory cell of the system contributes to the storage and recall of many (potentially all) stored items. Some of types of distributed memory have attractive properties of parallelism, resistance to noise and malfunctions, etc. However, exactly correct answers to the nearest neighbor queries from such memories are not guaranteed, especially when too many vectors are stored in the memory. Neurobiologically plausible variants of distributed memory can be represented as artificial neural networks. These typically perform one-shot memorization of vectors by a local learning rule modifying connection weights and retrieve a memory vector in response to a query vector by an iterative procedure of activity propagation between neurons via their connections.

In the first Section, we briefly introduce Hebb's theory of brain functioning based on cell assemblies because it has influenced many models of NAM. Then we introduce a generic scheme of NAMs and their characteristics (discussed in more details in the other sections). The following three Sections discuss the widespread matrix-type NAMs (where each pair of neurons is connected by two symmetric connections) of Hopfield, Willshaw, and Potts that work best with sparse binary vectors. The next Section is devoted to the function of generalization, which differs from the function of autoassociative memory and emerges in some NAMs. The following Section discusses NAMs with higher-order connections (more than two neurons have a connection) and NAMs without connections. Then some recent NAMs with a bipartite graph structure are considered. The last Section provides discussion and concludes the paper.

## CELL ASSEMBLIES AND GENERIC NAM

*Hebb's paradigm of cell assemblies.* According to Hebb [65], nerve cells of the brain are densely interconnected by excitatory connections, forming a neural network. Each neuron determines its membrane potential as the sum of other active neurons' outputs weighted by connection weights. A neuron becomes active if this potential (the input sum) exceeds the threshold value. During network functioning, connection weights between simultaneously active neurons (encoding various items) are increased (the Hebbian learning rule). This results in organization of neurons into cell assemblies — groups of nerve cells most often active together and consequently mutually excited by connection weights between neurons in the assembly. At the same time, the process of increased connection within assemblies leads to mutual segregation of assemblies. When a sufficient part of a cell assembly is activated, the assembly becomes active as a whole because of the strong excitatory connection weights between the cells within the assembly.

Cell assemblies may be regarded as memorized representations of items encoded by the distributed patterns of active neurons. The process of assembly activation by a fragment of the memorized item may be interpreted as the process of pattern completion or the associative retrieval of similar stored information when provided with a partial or distorted version of the memorized item.

Hebb's theory of brain functioning — interpretation of various mental phenomena in terms of cell assemblies — has turned out to be one of the most profound and generative approaches to brain modeling and has influenced the work of many researchers in the fields of artificial intelligence, cognitive psychology, modeling of neural structures, and neurophysiology (see also reviews in [39, 40, 54, 75, 98, 104, 120, 121, 134]).

*A generic scheme and characteristics of NAMs.* Let us introduce a generic model of the NAM type, inspired by Hebb's paradigm, that will be elaborated in the sections below devoted to specific NAMs. We mainly consider NAMs of the distributed and matrix-type, which are fully connected networks of binary neurons (but see Sections "NAMs with Higher-Order Connections and without Connections", "NAMs with a Bipartite Graph Structure for Nonbinary Data with Constraints" for other NAM types). Each of the neurons (their number is $D$) represents a component of the binary vector $\mathbf{z}$. That is, each of the $D$ neurons can be in the state 0 or 1. Each pair of neurons has two mutual connections (one in each direction). The elements of the connection matrix $\mathbf{W}(D \times D)$ represent the weights of all these connections. In the learning mode, the vectors $\mathbf{y}$ from the training or memory set (which we call the "base") are "stored" (encoded or memorized) in the matrix $\mathbf{W}$ by using some learning rule that changes the values of $w_{ij}$ (initially each $w_{ij}$ is usually zero).

In the retrieval mode, an input binary vector $\mathbf{x}$ (probe or key or query vector) is fed to the network by activation of its neurons: $\mathbf{z} = \mathbf{x}$. The input sum of the $i$-th neuron

$$s_i = \sum\nolimits_{j=1,D} w_{ij} z_j$$

is calculated. The neuron state is determined as

$$z_i(t+1) = 1 \text{ (active) for } s_i(t) \geq T_i(t)$$

and

$$z_i(t+1) = 0 \text{ (inactive) for } s_i(t) < T_i(t); \ T_i$$

is the value of the neuron threshold.

For parallel (synchronous) network dynamics, the input sums and the states of all $D$ neurons are calculated (updated) at each step $t$ of iterative retrieval. For sequential (asynchronous) dynamics, $z_i$ is calculated for one neuron $i$, selected randomly. For simplicity, let us consider random selection without replacement, and one step of the asynchronous dynamics to consist of update of the states of all $D$ neurons.

The parameters **W** and **T** are set so that after a single, or several, steps of dynamics the state of the network (neurons) reaches a stable state (typically, the state vector does not change with $t$, but cyclic state changes are also considered as "stable"). At the stable state, **z** is the output of the network.

The query vector **x** is usually a modified version of one of the stored vectors **y**. In the literature, this might be referred to as a noisy, corrupted, or distorted version of a vector. While the number of stored vectors is not too high, the output **z** is the stored **y** closest to **x** (in terms of dot product

$$\text{sim}_{\text{dot}}(\mathbf{x},\mathbf{y}) \equiv \langle \mathbf{x},\mathbf{y} \rangle ).$$

That is, **z** is the base vector **y** with the maximum value of $\langle \mathbf{x},\mathbf{y} \rangle$. In this case, NAM returns the (exact) nearest neighbor in terms of $\text{sim}_{\text{dot}}$. For binary vectors with the same number of unit (i.e. with value equal to 1) components, this is equivalent to the nearest neighbor by the Hamming distance ($\text{dist}_{\text{Ham}}$).

The time complexity (runtime) of one step of the network dynamics is $O(D^2)$. Thus, if a NAM can be constructed that stores a base of $N > D$ vectors so that they can be successfully retrieved from their distorted versions, then the retrieval time via the NAM could be less than the $O(DN)$ time required for linear search (i.e. the sequential comparison of all base vectors **y** to **x**). Since the memory complexity of this NAM type is $O(D^2)$, as $D$ increases, one can expect an increasing in the size $N$ of the bases that could be stored and retrieved by NAM.

Unfortunately, the vector at the NAM output may not be the nearest neighbor of the query vector, and possibly not even a vector of the base. (Note that if one was not concerned with biological plausibility, one can quickly check whether the output vector is in the base set by using a hash table to store all base vectors.) In some NAMs, it is only possible to store many fewer vectors $N$ than $D$, with high probability of accurate retrieval, especially if the query vectors are quite dissimilar to the base vectors.

For NAM analysis, base vectors are typically selected randomly independently from some distribution of binary vectors (e.g., vectors with the probability $p$ of 1-components equal to 1/2, or vector with $pD$ 1-components, for some $p$ from interval (0,1)). The assumption of independence simplifies analytical approaches, but is likely unrealistic for real applications of NAMs. The query vectors are typically generated by as modifications of the base vectors. Distortion by deletion randomly changes some of the 1s to 0s (the remaining components are guaranteed to agree). A more complex distortion by noise randomly changes some 1s to 0s, and some 0s to 1s while (exactly or approximately) preserving the total number of 1s.

For a random binary vector of dimension $D$ with the probability $p$ of a component to be 1, the Shannon entropy

$$H = Dh(p),$$

Where

$$h(p) = -p \log p - (1-p) \log(1-p).$$

For $D \gg 1$, a random vector with $pD$ of 1s has approximately the same entropy. The entropy of $N$ vectors is $NDh(p)$. When $N$ vectors are stored in NAM, the entropy per connection is [40, 41].

$$\alpha = NDh(p)/D^2 = Nh(p)/D.$$

Knowing $h(p)$, it is easy to determine $N$ for a given $\alpha$.

When too many vectors are stored, NAM becomes overloaded and the probability of accurate retrieval drops (even to 0). The value of $\alpha$ for which a NAM still works reliably depends on the mode of its use (in addition to the NAM design and distributions of base and query vectors). The mode where the undistorted stored base vectors are still stable NAM states (or stable states differ by few components from the intended base vectors), has the largest value of $\alpha$. (We denote the largest value of $\alpha$ for this mode as "critical", $\alpha_{crit}$, and the corresponding $N$ as $N_{crit}$.) For $\alpha > \alpha_{crit}$ the stored vectors become unstable. Note that checking if an input vector is stable does not allow one to extract information from the NAM, since vectors not stored can also be stable.

The information (in the Shannon information-theoretic sense) that can be extracted from a NAM is determined by the information efficiency (per connection) $E$. This quantity is bounded above by some specific $\alpha$ (the entropy per connection that still permits information extraction), which in its turn is bounded above by $\alpha_{crit}$. A NAM may work in recognition or correction mode. In recognition mode, the NAM distinguishes whether the input (query) vector is from the base or not, yielding extracted information quantified by $E_{recog}$.

When NAM answers the nearest neighbor queries (correction mode), information quantified by $E_{corr}$ is extracted from the NAM by correction (completion) of the distorted query vectors. The more distorted the base vector used as the query at the NAM input, the more information $E_{corr}$ is extracted from the NAM (provided that the intended base vector is sufficiently accurately retrieved). However, more distorted input vectors lower the value of $\alpha_{corr}$ at which the NAM is still able to retrieve the correct base vectors, and so lowers $E_{corr}$ (which is constrained to be less than $\alpha_{corr}$). We refer to these information-theoretic properties of NAMs as their information characteristics.

Let us now consider specific variants of the generic NAM. Hereafter we use the terms "NAM" and "(neural) network" interchangeably.

## HOPFIELD NAMs

***Hopfield networks with dense vectors.*** In the Hopfield NAM, "dense" random binary vectors (with the components from {0,1} with the probability $p = 1/2$ of 1) are used [68]. The learning procedure forms a symmetric matrix **W** of connection weights with positive and negative elements. The connection matrix is constructed by successively storing each of the base vectors **y** according to the Hopfield learning rule:

$$w_{ij} = w_{ij} + (y_i - q)(y_j - q)$$

with parameter $q = p = 1/2$; $w_{ii} = 0$. (For brevity, we use the same name for generalization of this rule with $q < 1/2$, though Hopfield did not propose it, Subsection "Hopfield networks with sparse vectors").

The dynamics in [68] is sequential (in many subsequent studies and implementations it is parallel) with the threshold $T = 0$. It was shown [68] that each neuron state update decreases the energy function

$$-(1/2)\sum_{i,j=1,D} z_i w_{ij} z_j ,$$

so that a (local) minimum of energy is eventually reached and such a network comes into a stable activation state.

As $D \to \infty$, various methods of analysis and approximation of experimental (modeling) data obtain $\alpha_{crit} \approx 0.14$ [68, 8, 5, 71, 35] which gives $N_{crit} \approx 0.14D$ since $h(1/2) = 1$. Note that similar values of $\alpha_{crit}$ are achieved at rather small finite $D$. For rigorous proofs of (smaller) $\alpha_{crit}$ see refs in [107].

As for $E_{corr}^{max}$ for distortion by noise, 0.092 was obtained by the method of approximate dynamical equations of the mean field [71], and 0.095 by approximating the experiments to $D \to \infty$ [35].

By the coding theory methods in [112] it was shown that asymptotically (as $D \to \infty$) it is possible to retrieve (with probability approaching 1) exact base vectors with query vectors distorted by noise (so that their $\text{dist}_{Ham} < D/2$ from the base vectors), for $N = D/(2\ln D)$ stored base vectors if non-retrieval of some is permitted. If one requires the exact retrieval of all stored base vectors, the maximum number of vectors which can be stored decreases to $N = D/(4\ln D)$. These values of $N$ were shown to be the lower and upper bounds in [25, 20]. Note that in [47] $\alpha_{crit} = 2$ was obtained for "optimal" **W** (obtained by a non-Hebbian learning rule); a pseudoinverse rule (e.g. [125, 140]) gives $\alpha_{crit} = 1$.

For correlated base vectors, the storage capacity $N_{crit}$ depends on the structure of the correlation. When the base vectors are generated by a one-dimensional Markov chain [107], $N_{crit}$ is somewhat higher than it is for independent vectors. This and other correlation models were considered in [108].

***Hopfield networks with sparse vectors.*** Hopfield NAMs operating with sparse vectors $p < 1/2$ appeared to have better information characteristics [154] (see also Sections "Willshaw NAMs", "Potts NAMs") than those operating with dense vectors ($p = 1/2$). For example, they attain values $N > D$.

In the usual Hopfield NAM and learning rule (with $q = p = 1/2$ and threshold $T = 0$) the number of active neurons is kept near $D/2$ by the balance of negative and positive connections in $\mathbf{W}$. Using the Hopfield rule with $q = p < 1/2$ one can not set $T = 0$. This is especially evident for the Hebb learning rule (which we obtain from the Hopfield rule by setting $q = 0$). All connections become non-negative, and $T = 0$ eventually activates all neurons. Similar behavior is demonstrated by the Hopfield NAM and learning rule with $q = p < 1/2$. The problem of network activity control (i.e. maintaining some average activity level chosen by the designer) can be solved by applying an appropriate uniform activation bias to all neurons [9, 21]. This is achieved by setting an appropriate positive value of the time-varying threshold $T(t)$ [21] to ensure, for example, $pD < D/2$ active neurons (to match $pD$ in the stored vectors) for parallel dynamics.

Note that the Hopfield rule with $q = p < 1/2$ provides better information characteristics than the pure Hebb rule with $q = 0$ [41, 35]. However, the Hebb rule requires modification of only $(pD)^2$ connections per vector, whereas the Hopfield rule modifies all connections per vector.

As $D \to \infty$ and $p \to 0$ ($pD \gg 1$, and often $p \sim \ln D / D$) the theoretical analysis (e.g., [154, 41, 34] and others) gives

$$\alpha_{\text{crit}} = (\log e)/2 = 1/(2\ln 2) \approx 0.72$$

for the Hopfield rule with $q = p$, the Hebb rule, and the "optimal" $\mathbf{W}$ [47]. In [34] they use a scaled sparseness parameter $\varepsilon = (\ln p)^{-1/2}$ to investigate convergence of $\alpha_{\text{crit}}$ to $\alpha_{\text{crit}}^{\max}$. For $\varepsilon \ll 1$ they obtained $\alpha_{\text{crit}}^{\max} \approx 0.72$. However for $\varepsilon = 0.1$ (corresponding to $p = 10^{-8}$ and to $D > 10^9$), $\alpha_{\text{crit}} = 0.43$ only.

In [122] it was shown that $E_{\text{recog}}^{\max} = 1/(4\ln 2) \approx 0.36$ (by the impractical exhaustive enumeration procedure of checking that all vectors of the base are stable and all other vectors with the same number of 1-components are not stable). This empirical estimate coincides with the estimate [41]. For retrieval by a single step of dynamics,

$$E_{\text{corr}}^{\max} = 1/(8\ln 2) \approx 0.18$$

for distortion by deletion of half the 1-components of a base vector [40, 41, 119, 146].

Let us note again, all these results are obtained for $D \to \infty$ and $p \to 0$. For these conditions, multi-step retrieval (the usual mode of NAM retrieval as explained in Subsection "A generic scheme and characteristics of NAMs") is not required since NAM reaches a stable state after a single step. In terms of $N$, since $h(p) \to 0$ for $p \to 0$, it follows that $N >> D$, that is

$$N_{\text{crit}} = \alpha_{\text{crit}} D / h(p) \to \infty$$

much faster than $D \to \infty$.

The same is valid for $N$ corresponding to $E$.

In the experiments [146], for the Hebb rule and multi-step retrieval $E_{\text{corr}}$ values up to 0.09 were obtained. Detailed studies of the information characteristics of the finite- and infinite-dimensional networks with the Hopfield rule, can be found in [34, 35]. Different degrees of distortion by noise for vectors with $pD$ unit components were used. The dynamic threshold ensured $pD$ active neurons at each step of the parallel dynamics. It was shown [35] that with this choice of threshold the stable states are static (some vector) or cycles of length 2 (two alternating vectors on adjacent steps of the dynamics). (This is the same behavior as for the fixed static threshold and is valid for all networks with symmetric connections.) It has been demonstrated experimentally [35] that even if after the first step of dynamics

$$\text{sim}_{\text{dot}}(\mathbf{y}, \mathbf{z}) < \text{sim}_{\text{dot}}(\mathbf{y}, \mathbf{x})$$

(where $\mathbf{y}$ is the correct base vector, $\mathbf{z}$ is the network state, and $\mathbf{x}$ is the distorted input), the correct base vector can sometimes be retrieved by the subsequent steps of the dynamics. Conversely, increasing $\text{sim}_{\text{dot}}(\mathbf{y}, \mathbf{z})$ at the first step of the dynamics does not guarantee correct retrieval [5]. These results apply to both the dense and sparse vector cases. The study [35] used analytical methods developed for the dense Hopfield network and adapted for sparse vectors, including the statistical neurodynamics (SN) [5, 34], the replica method (RM) [8], and the single step method (SS) [80].

All these analytical methods rather poorly predicted the behavior of finite networks for highly sparse vectors, at least for the parallel dynamics studied. (Note that all these methods (SN, RM, SS) provide accurate results for $D \to \infty$ and $p \to 0$, where retrieval by a single step of dynamics is enough.) Empirical experimentation avoids the shortcomings of these analytical methods by directly simulating the behavior of the networks. These simulations allow $\alpha_{\text{corr}}$ and information efficiency, $E_{\text{corr}}$, to be estimated as a functions of $p, D$ and the level of noise in the input vectors. The value of $E_{\text{corr}}$ monotonically increases as $D$ increases for a constant $p$. For $p = 0.001 - 0.01$, which corresponds to the activity level of neurons in the brain, the maximum value of $E_{\text{corr}} \approx 0.205$ was obtained by approximating experimental results to the case $D \to \infty$ [35] (higher than $E_{\text{corr}}^{\max} \approx 0.18$ for $p \to 0$).

In [38] the time of retrieval in the Hopfield network was investigated (using the number of retrieval steps observed in simulation experiments; this number somewhat increases with $D$). They conclude that for random vectors with small $p$, large $D$, and large $N$, Hopfield networks may be faster than some other algorithms (e.g., the inverted index) for approximate and exact nearest neighbor querying.

An increase in the number $N_{\text{crit}}$ of stable states corresponding to the stored vectors proportional to $(D/\ln D)^2$ for $p \sim \ln D / D$ is shown asymptotically in [4] (although non-rigorously, see also [34]). Note, this result also follows from $N = \alpha D / h(p)$ by approximating $h(p) \approx -p \log p$ for small enough $p$ ($-\ln p \gg 1$).

In [67] they give a rigorous analysis of a Hopfield network variant (neurons are divided into parts, see Section "Potts NAMs"), with the Hebb learning rule and $p$ slightly less than $\ln D / D$, for retrieval by a single step of parallel dynamics with fixed $T$. The lower and upper bounds of $N$ were obtained for which the memory vectors are stable states (with probability approaching 1 as $D \to \infty$), and can also be exactly retrieved from query vectors distorted by noise. The lower and upper bounds of $N$ found in [67] are of the same order as those found in [4]. For this mode of network operation, if we approximate the number of retrieval steps as $\ln D$, we may estimate speed-up as $D/(\ln D)^3$ relative to linear search (see Subsection "A generic scheme and characteristics of NAMs").

For both dense and sparse vectors, the NAM capacity $N^{\max}$ grows with increasing $D$. Also, in order to maintain an adequate information content for a sparse vector ($Dh(p)$ for $h(p) \ll 1$), it is necessary to have a sufficiently high $D$. The number of connections grows as $D$ squared (because Hopfield networks are fully connected), which is unattainable even on modern computers at $D$ of millions. Besides, the neurobiologically plausible number of connections per neuron is on the order of 10,000. Therefore, the development of "diluted" networks that perform NAM functions without being fully connected is attractive, e.g. [105, 150, 151, 41, 142]. This partial connectivity can be used to reduce the memory complexity of NAM from quadratic to linear in $D$ [98, 99].

## WILLSHAW NAMs

***Willshaw networks with sparse vectors.*** NAMs with binary connections from {0,1} are promising since they require only one bit per connection. Such networks were proposed both in heteroassociative [157] and autoassociative versions (e.g. [118, 156, 16, 152, 49, 115, 119, 122, 24, 48, 41, 42, 43, 44, 81]). The learning rule (let's call it the Willshaw rule) becomes:

$$w_{ij} = w_{ij} \vee (y_i \wedge y_j),$$

where $\vee$ is disjunction, $\wedge$ is conjunction. Various strategies for threshold setting can be used, e.g., setting threshold $T$ to ensure $pD$ active neurons, as in Subsection "Hopfield networks with sparse vectors".

Note that this NAM can not work with dense vectors, since storing only a small number of dense vectors will set almost all the connection weights to 1. Moreover, for the same reason, the Willshaw networks (unlike the Hopfield networks) cease to work at any constant $p$ and $\alpha$ as $D \to \infty$. The number $N$ of random binary vectors able to be stored and retrieved in the Willshaw NAM grows with decreasing vector density. Even for not very large networks and not very sparse vectors ($p \sim \sqrt{D}/D$), $N$ can exceed $D$ (e.g. $D = 4096$ allowed storage and retrieval of up to $N = 7000$ vectors distorted by noise in the experiments of [16]). The particular $N$ values reached (for $D$ and $p$ fixed) vary depending on the degree of the query vector distortion and on the desired probability of retrieval.

The maximum theoretical $\alpha_{\text{crit}} = \ln 2 \approx 0.69$ is reached as $D \to \infty$ for $p \sim \ln D / D$ [40, 119, 41]. In [122] they obtained

$$E_{\text{recog}}^{\max} = \ln 2 / 2 \approx 0.346$$

(using a computationally expensive exhaustive enumeration procedure). In [42, 43] the same $\alpha_{\text{crit}}$ and $E_{\text{recog}}^{\max}$ were obtained analytically for the sparseness parameter $\beta = \log(1/p)/(pD)$ equal to 1, i.e. for $pD$ somewhat less than $\log D$. (The probability of a connection to be modified after storing $N$ vectors is $1-(1-p^2)^N \approx 1-\exp(-Np^2) = 1-\exp(-\alpha D p^2 / h(p)) = 1-\exp(-\alpha p \log(1/p)/(\beta h(p))) \approx 1-\exp(-\alpha/\beta) < 1$ (we used $p \to 0$); thus the network can be analyzed for fixed $\alpha$ and $\beta$ at $D \to \infty$.) The same upper bound of $E$ is given as the maximum entropy of **W** learnt by the Willshaw rule. In [119] the efficiency

$$E_{\text{corr}}^{\max} = \ln 2 / 4 \approx 0.173$$

was theoretically shown for single-step (as well as multi-step) retrieval and distortion by deletion.

For multi-step retrieval in finite Willshaw NAMs (with distortion by deletion) $E_{\text{corr}}$ up to 0.19 (at $D = 20000$) was obtained experimentally in [146]. Experiments in [146, 44] show that in the Willshaw NAM (unlike the Hopfield NAM), the values $\alpha_{\text{corr}}$ and $E_{\text{corr}}$ for not too large $D$ are higher than for $D \to \infty$ (see also [42]). Note that the quality of retrieval in the Willshaw NAM is higher than in the Hopfield NAM; the retrieved vectors more often coincide exactly with the stored vectors of the base.

From the detailed analytical and experimental study of the values of $\alpha_{\text{corr}}$ in [44] (at various levels of sparsely, parameterized as $\beta$, degrees of distortion by noise, and $D$ up to 100000), it was found that $E_{\text{corr}} \approx 0.13$ per connection can be reached in the experiments (for small networks, $N = 640$, $pD = 20$, $\beta = 0.25$). It was also shown that the results of the analytical methods SS [80] and GR [48] are far from the experimental results (in most cases, worse than them). Due to the connections being binary, the efficiency per bit of connection implemented in computer memory is higher than that for the Hopfield network (where $E_{\text{corr}} \approx 0.205$ per connection [41]).

A review of NAM studies in [81] concludes that for Willshaw networks having connection matrix **W** with probability of a nonzero element close to 0 or 1, compression of **W** improves information characteristics compared with the usual uncompressed Willshaw NAM. Such compressed **W** are obtained when the base vectors have the number of 1s sublogarithmic or superlogarithmic in $D$. Their comparison of the retrieval time in compressed Willshaw networks and the inverted index has shown the advantage of the inverted index for most parameters.

An analytical and experimental comparison of the Willshaw, the GB (Subsection "Willshaw-Potts network"), and the Hopfield networks (with the Hebb rule [4]) for vectors with $p$ of the order of $\ln D / D$ and distortion by deletion was carried out in [59]. They investigate single-step retrieval theoretically (asymptotically, for $D \to \infty$ with probability approaching 1). For all models, the lower bound of $N$ of the order $(D / \ln D)^2$ is obtained, and for the Willshaw network the matching upper bound is shown. In experiments, the results are worse for a fixed threshold than for a variable threshold. The Hopfield network performed worse, in terms of empirical probability of retrieval versus $N$, compared with the other NAMs, probably because of the non-optimal Hebb learning rule and non-optimal threshold selection.

For the diluted Willshaw networks [24, 6, 41, 99] the optimal $pD$ (providing approximately the same capacity $N$) is higher than for the fully connected networks.

***Willshaw networks in the index structures for nearest neighbor search***. In [159] the base of binary sparse vectors is divided into disjoint sets (of the same cardinality) and each is stored in a Willshaw NAM with its own **W**. When the query vector **x** is input, $\text{sim}_{\text{dot}}(\mathbf{x}, \mathbf{Wx})$ is calculated for the matrices **W** of all sets, and the vectors of sets with the maximum similarity are used as the nearest neighbor candidates (verified by linear search). Analysis and experiments for bases of random vectors with small random distortions of query vectors showed that up to a certain number of vectors in each network the nearest neighbor is found with a high probability (in experiments, without error) and faster than by linear search only. If this number of vectors per network is exceeded, both the probability of finding an incorrect nearest neighbor and its distance to the correct vector increase. A somewhat lower speedup relative to linear search is shown for real, nonrandom data, versus synthetic, random data. In [60] similar results were obtained analytically (asymptotically for $D \to \infty$ and error probability approaching zero) and experimentally for bases of sparse and dense random vectors (for the Hopfield rule).

## POTTS NAMs

***Potts networks.*** The NAM from [77] can be considered as a network of neurons that are divided into non-overlapping parts ("columns"), with $d$ neurons in each column and only one active neuron in the state $z = d - 1$, for the remaining column neurons $z = -1$. That is, the sum of activations over all the neurons is zero in each column. The Hebb rule is used for learning.

For the more convenient version of this model with the neurons having the states from {0,1} and single active neuron per column, the Hopfield rule is used. The connection matrix **W** for the entire network is constructed so that $w_{ij} = 0$ for neurons $i$ and $j$ in the same column (this implies that $w_{ii} = 0$). That is, the network is structured as a multipartite graph. Network dynamics (parallel or sequential) activates the one neuron in each column with the maximum input sum $s$ (one of these neurons is randomly activated for neurons with equal $s$, but see the GB network below).

For the number of columns $D$, the value of $N_{\text{crit}}/D$ of the Potts network was estimated by [77] to be $d(d-1)/2$ times more than 0.14 (i.e., $\alpha_{\text{crit}}$ for the Hopfield network with $p=1/2$). However, to approximate the number of connections in the Hopfield network, the Potts network must have $D/d$ columns. Note also that one "Potts vector" contains only $(D/d)\log d$ bits of information [79].

The Potts network with parallel and sequential dynamics, and with single-step and multi-step retrieval was analytically explored in [109]. For exact retrieval (asymptotically, as $D \to \infty$, with probability approaching 1), the upper and lower bounds of $N$ were estimated both for the mode of querying with stable stored vectors and for the correction mode querying with distorted query vectors. In both cases,

$$N = cD/\ln D,$$

where the constant $c$ increases quadratically in $d$, but with different $c$ depending on the degree of distortion and the desired probability of state to be stable or vector to be retrieved.

***Willshaw-Potts network.*** For binary connections with the Willshaw learning rule, the Potts network becomes the Willshaw-Potts network [79]. When a vector is stored, a clique (a complete subgraph) is created in the connection graph. As for the Hopfield network, only static stable states or cycles of length 2 were experimentally observed for parallel dynamics. According to [79], the information characteristics of this network are close to the Willshaw network at the same vector sparsity. Since the information content of Willshaw-Potts vectors is low, the $N_{\text{crit}}$ is higher than for the Willshaw network.

This network was rediscovered as the GB network in [58] with various modifications [3, 59] and hardware implementations (for example, [117] with non-binary connections). The GB network is oriented for exact retrieval of vectors with distortion by deletion (columns without values activate all neurons). The peculiarities of the GB network include: connections of neurons with themselves; the possibility for several neurons in a column (with the maximum input sum) to be in the active state; the contribution from each column to the input sum of a neuron is not more than 1; various options for threshold management; the possibility of working with vectors having all zero components in some columns, etc. A theoretical GB analysis for single-step retrieval, as well as an experimental comparison with the Hopfield and the Willshaw networks for multi-step retrieval is given in [59], see also Subsection "Willshaw networks with sparse vectors".

***Processing of realistic data.*** To represent arbitrary binary vectors in the Potts network, they are divided into segments of dimension $\log d$ and each segment is encoded by the activation of one neuron of its $d$-dimensional column [97]. Components of integer vectors can be represented in a similar way.

Simulated data are typically generated as independent random samples. This ensures that the vectors to be stored are very nearly equally and maximally dissimilar. However, data generated from situations in the world is very unlikely to be so neatly distributed. When working with (unevenly distributed) real data, NAM is used non-optimally (many connections are not modified, others are "oversaturated"). To overcome this in the GB network, a free column neuron is allocated when the number of connections of a neuron (encoding some value) exceeds the threshold [19]. During retrieval, all column neurons that encode a certain value are activated.

For better balancing the number of connections, in [64] the number of neurons in the column allocated to represent a vector component is proportional to the frequency of its 1-value in the vectors of the base. During storage, the various neurons representing the component are activated in turn. The authors of [64] also propose an algorithm for finding (with a high probability) all vectors of the base closest to the query vector distorted by deletion; the algorithm often significantly reduces the number of required queries.

## GENERALIZATION IN NAMs

The Hebbian learning in matrix-type distributed memories naturally builds a kind of correlation matrix where the frequencies of joint occurrence of active neurons are accumulated in the updated connection weights. The neural assemblies thus formed in the network may have a complex internal structure reflecting the similarity structure of stored data. This structure can be revealed as stable states of the network — in the general case, different from the stored data vectors. That is, it is possible for vectors retrieved from a network to not be identical to any of the vectors stored in the network (which is generally undesirable).

***Similarity preserving binary vectors.*** Similarity of patterns of active neurons (represented as binary vectors) are assumed to reflect the similarity of items (of various complexity and generality) they encode. The similarity value is measured in terms of the number or fraction of common active neurons (or overlap, i.e. normalized dot product of the representing binary vectors). Moreover, the similarity "content" is available as the identities of common active neurons (the IDs of the common 1-components of the representing binary vectors).

Note that such data representation schemes by similarity preserving binary vectors have been developed for objects represented by various data types, mainly for (feature) vectors (see survey in [131]), but also for structured data types such as sequences [102, 72, 85,86] and graphs [127, 128, 148, 136, 62, 134]. A significant part of this research is developed in the framework of distributed representations [45, 76, 106, 126, 89], including binary sparse distributed representations [102, 98, 103, 127, 128, 113, 114, 137, 138, 139, 148, 135, 136, 61, 134, 129, 130, 131, 132, 31, 33] and dense distributed representations [75, 76] (see [82, 84, 87, 88, 83] for examples of their applications).

***Complex internal structure of cell assemblies for graded connections.*** When binary vectors reflecting similarity of real objects are stored in a NAM by variants of the Hebb (or Hopfield) rule, the weights of connections between the neurons frequently activated together will be greater than the mean value of all the weights. On the other hand, rare combinations of active neurons will have smaller weights. Thus, neuron assemblies (cell assemblies in terms of Hebb [65]) formed in the network may have a very complicated structure. Hebb and Milner introduced the notions of "cores" and "fringes" ([65] pp. 126–134; for more recent research see [101]) to characterize qualitatively such complex internal structure of assemblies.

The notions of core (kernel, nucleus) and fringe (halo) of assemblies have attracted attention to the function of assemblies distinct from the function of associative memory. This different function is not just memorization of individual activity patterns (vectors), but emergence of some generalized internal representations, that were not explicitly presented to the network as vectors for memorization.

Some assembly cores may correspond to "prototypes" representing subsets of attributes (encoded by the active neurons represented by the 1-components of the corresponding vectors) often present together in some input vectors. Note that some of these subsets of components/attributes may never be present in any single vector. Stronger cores, corresponding to stable combinations of a small number of typical attributes present in many vectors employed for learning, may correspond to some more abstract or general object category (class). Cores formed by more attributes may represent more specific categories, or object prototypes. However, object tokens (category instances) may also have strong cores if they were often presented to the network for learning. Note that some mechanisms may exist to prevent repeated learning of vectors that are already "familiar" to the network. Also, the rate of weight modification may vary based on the "importance" of the input vector.

The representations of real objects have different degrees of similarity with each other. Similarities in various combinations of features often form different hierarchies of similarities that reflect hierarchies of categories of different degrees of generalization (object — class of objects — a more general class, etc.). So, assemblies formed in the network (cores and fringes of different "strength") may have a complex and rich hierarchical structure, with multiple overlapping hierarchies reflecting the structure of different contents and values of similarity implicitly present in the base of vectors used for the unsupervised network learning by the employed variant of the Hebb rule. Thus, many types of the category-based hierarchies (also known as generalization or classification or type-token hierarchies) may naturally emerge in the internal structure of assemblies formed in a single assembly neural network (NAM).

Complex internal structure of a neural assembly allows a virtually continuous variety of hierarchical transitions. To reveal various types of categories and prototypes and instances formed in the network, the corresponding assemblies should be activated. To activate only stronger cores, higher values of threshold should be used. Lower threshold values may additionally activate fringes.

***Research of generalization function in NAMs.*** Additional stable states that emerge in NAM after memorizing random base vectors and do not coincide with the base vectors are known as false or spurious states or memories, e.g., [69, 155]. In [155] they regarded the emergence of spurious attractors in the Hopfield networks as a side effect of the main function of distributed NAMs, consisting not in memorizing individual patterns, but in formation of prototypes. Such an interpretation is close to the earlier work [13, 11, 12] that considered formation of concepts, prototypes, and taxonomic hierarchies as a natural generalization of correlated patterns memorized in a distributed memory.

Research of hierarchically correlated patterns and states in Hopfield networks has been initiated by physicists who studied the "ultrametric" organization of ground states in spin-glasses (e.g., [123, 32, 63]; see also [6] and its references). While these earlier works required explicit representation of patterns at various hierarchical levels to be used in the learning rule, more neurobiologically plausible and practical Hebb and Hopfield rules applied to (hierarchically) correlated sparse binary patterns themselves (obtained with some simple correlation model) were considered, e.g., in [153, 66]. [153] theoretically showed "natural" formation of stable cores and fringes as well as traveling through different levels of hierarchies by uniform changing of the threshold. More complex probabilistic neuron dynamics and threshold control expressing neuronal fatigue was modeled in [66]. Dynamics of transitions between stable memory states that models human free recall data and can also be used with hierarchically organized data was considered in [143]. The "neuro-window" approach of [74] may be considered as using multiple thresholds to activate cores or fringes. Revealing the stable states corresponding to emergent assemblies is used for data mining (binary factor analysis) in [36, 37].

***Generalization in NAMS with binary connections.*** The Willshaw learning rule does not form assemblies with the complex internal structure needed for generalization functions, such as emergence of generalization (type-token) hierarchies. The Willshaw learning rule causes the connectivity of an assembly (corresponding to a vector) to become full after a single learning act (vector storage) and not change thereafter. To preserve the capability of forming assemblies with a non-uniform connectivity in NAMs with binary connections, a stochastic analogue of the Hebbian learning rule for binary connections was proposed in [100, 98]:

$$w_{ij} = w_{ij} \vee (y_i \wedge y_j \wedge \xi_{ij}),$$

where $\xi_{ij}$ is a binary random variable equal to 1 with the probability that determines the learning rate.

The connectivity value for some set of neurons is determined here by the number of their 1-weight connections. Neurons that have more than some fraction of 1-weight connections with the other neurons of the same assembly may be attributed to the core part of the assembly.

In [15] they experimentally studied formation of assemblies with cores and fringes using the above mentioned "stochastic Willshaw" rule ($D = 4096$, $pD = 120 - 200$, about 60 neurons in the core and 60–140 neurons in the fringe). Tests have been performed on retrieving a core by its part; a core and its full fringe by the core and a part of the fringe (the most difficult test); and, a core by a part of its fringe. As expected, experiments with correlated base vectors have shown a substantial decrease of storage capacity compared to random independent vectors. A special learning rule was proposed to increase the stability of fringes.

Formation of prototypes with the stochastic Willshaw rule was also investigated in [7, 23]; a model of paired-associate learning in humans is considered in [141].

***Generalization in modular NAMs.*** A modular structure of neural networks where the Hebb assemblies are formed inside the modules was proposed and developed in [50–56]. The modular assembly neural network is intended for recognition of a limited number of classes. The network is artificially partitioned into several modules (sub-networks) according to the number of classes that the network is required to recognize. Each module network is full-connected, connections are graded. The features extracted from all objects of a certain class are encoded into activation of the patterns of neurons within the corresponding sub-network. After learning, the Hebb assemblies are formed in each module network. In this modular structure, the network acquires the capability to generalize the description of each class within the corresponding module (sub-network), i.e. separately and completely independently from all other classes. In [56] it was shown that the number of connections in each module can be reduced without loss of the recognition capability.

## NAMs WITH HIGHER-ORDER CONNECTIONS AND WITHOUT CONNECTIONS

Neural networks in the previous sections have connections of order $n = 2$ (a connection is between two neurons). In this section we consider values of $n$ other than 2, for the NAMs with the structure of the Hopfield network (unlike Section "Hopfield NAMs" where we only considered the case $n = 2$).

***Neural networks with higher-order connections.*** In the higher-order (order $n > 2$) generalization of the Hopfield network, $n$ neurons are connected by single connection instead of just two (for example, [124, 17, 46, 1, 70, 94, 26, 95, 96, 30]). For the neuron with states from $\{-1, +1\}$ ($p = 1/2$), the network dynamics can be defined as

$$z_i = \mathrm{sign}(\sum_{j_1 \ldots j_n \neq i} w_{i j_1 \ldots j_n} z_{j_1 \ldots j_n}).$$

The analogue of the Hebb learning rule becomes

$$w_{i_1 \ldots i_n} = 1/D^{n-1} \sum_{\mu=1,N} y^{\mu}_{i_1} y^{\mu}_{i_2} \ldots y^{\mu}_{i_n}.$$

Other learning rules can also be used.

The number of stable states corresponding to the stored random binary vectors (possibly slightly different from them) is estimated in the mentioned papers to be $N_{\mathrm{crit}} \approx \alpha_{crit}(n) D^{n-1}$. As in NAMs with connections between pairs of neurons, $\alpha_{\mathrm{crit}}$ depends on the specific type of learning rule and network dynamics. $\alpha_{\mathrm{crit}}$ does not exceed 2 and decreases with increasing $n$ [94]. For the absence of errors (with a probability approaching 1), the number of stored vectors

$$N \approx 1/c_n \, D^{n-1} / \ln D$$

(for example, [46, 26, 95, 30]). In [95] they obtain $c_n > 2(2n-3)!!$ So, the exponential in $n$ growth of $N$ is due to the exponential growth of the connection number, and the characteristics per connection deteriorate with increasing $n$.

***The generalization of Krotov-Hopfield.*** For networks with higher-order connections, the network energy in [95] is written as

$$-\sum_{\mu=1,N} F(\langle \mathbf{z}, \mathbf{y}^\mu \rangle)$$

with a smooth function $F(u)$. For polynomial $F(u)$ and $n=2$, this gives the energy of the usual Hopfield network ([68] and Subsection "Hopfield networks with dense vectors"). For small $n$, many memory vectors $\mathbf{y}^\mu$ have approximately the same values of $F(u)$ and make a comparable contribution to the energy. For $n \to \infty$, the main contribution to the energy is given by the memory vector $\mathbf{y}$ with the largest $\langle \mathbf{z}, \mathbf{y} \rangle$. For intermediate $n$, a large contribution is made by several nearest memory vectors.

In [30] it is proved that for $F(u) = \exp(u)$ this memory allows one to retrieve $N = \exp(\alpha D)$ randomly distorted vectors (within $\text{dist}_{\text{Ham}} < D/2$ from the stored vectors) by a single step of the sequential dynamics, for some $0 < \alpha < \ln 2/2$, depending on the distortion, with probability converging to 1 for $D \to \infty$.

In [95] they consider the operation of such a network in the classification mode, where each stored base vector corresponds to one of the categories to be recognized. In particular, to classify the handwritten digits of the MNIST base into 10 classes, in addition to the "visible" neurons to which 28x28 images (with pixel values in [−1, + 1]) are input, there are 10 "classification" neurons. The value of the output is obtained by a non-linearity $g(s)$ applied to the input sum $s$, for example, $\tanh(s)$ (instead of the $\text{sign}(s)$ function used in the memory mode). The outputs of visible neurons are fixed, and the outputs of classification neurons are determined by a single step of the dynamics.

Vectors of $N$ memory states are formed by learning on the training set. The $N = 2000$ memory vectors minimizing the classification error for the 60,000 images of the MNIST training set were obtained with the stochastic gradient descent algorithm.

For a single step of the dynamics this structure is equivalent to a perceptron with one layer of $N$ hidden neurons [95]. The nonlinearity at the output of the visible neurons is $f(u) = F(u)$, and that at the output of hidden neurons is $g(u)$. The learned memory vectors (with components normalized to [−1, + 1]) are encoded in the weight vectors of the connections between the visible neurons and the hidden neurons. It is shown that when $n$ changes the visualized memory vectors change. For small $n$, the memorized vectors correspond to the features of the digit images, and for large $n$ they become prototypes of individual digits [95, 96].

***Neural networks with first-order connections.*** In a neural network with connection order $n=1$, each neuron is connected only with itself. They can be considered as networks without connections, where learning changes the state of the neurons themselves (with "neuron plasticity" [39]). Thus, memory is a single vector of the dimension of the vectors of the base.

For binary connections, we get the Bloom filter (see the reviews [22, 149]), which exactly recognizes the absence of an undistorted query vector in the stored database by absence of at least one of its 1-components in the memory vector. If the 1-components of the query vector are a subset of 1-components of the memory vector, the vector is recognized as the base vector, but it is necessary to check this, since there is a false positive probability due to "ghosts" (vectors not from the base, the 1-components of which belong to the memory vector). Ghosts can be considered as analogous to spurious memories (Subsection "Research of generalization function in NAMs").

An analysis of the probability of their appearance under certain restrictions on stored random vectors is given in [144]. In [158], they reduce the probability of false positives. In [57], a Bloom filter version is analyzed which recognizes the absence of distorted query vectors. The autoscaling Bloom filter approach proposed in [92] suggests a generalization of the counting Bloom filter approach based on the mathematics of sparse hyperdimensional computing and allows elastic adjustment of its capacity with probabilistic bounds on false positives and true positives. In [90], the formation of sparse memory vectors (with an additional operation of context-dependent thinning [134]) is considered, and in [91] the probability of correct recognition is estimated. The use of graded connections (the formation of the memory vector is done by addition), including subsequent binarization, and the classification problem for vectors not from the base, are considered in [89, 91].

For real-valued vectors and connections, the recognition of random undistorted vectors is analyzed in [10, 126]. In [73] they allow distortion of vectors. In [126, 73], the analysis of non-random base vectors is given.

## NAMs WITH A BIPARTITE GRAPH STRUCTURE FOR NONBINARY DATA WITH CONSTRAINTS

In some recent papers (e.g., [78, 145, 110, 111]), in order to create NAMs which can store and retrieve (from rather noisy input vectors) the number $N$ of (not always binary) vectors with $N$ near exponential in $D$, the vectors considered are not arbitrary random but satisfy (linear) constraints. The neural network has the structure of a bipartite graph. One set of neurons (not connected with each other) is used to represent the vectors of the base, neurons of the other set represent constraints. A rectangular matrix of connections between these two sets is learned on the vectors of the base. The connection vector of each constraint neuron represents the vector of that particular constraint. Iterative algorithms with local neuron computations are used for retrieval.

*Iterative algorithms for learning constraint matrix and vector recovery*. In [78, 145], they consider the problem of the exact retrieval (with high probability) of vectors that belong to a subspace of dimension less than $D$. The graded weights of the bipartite graph connections representing linear constraints are learned from the vectors of the base (which have only non-negative integer components). Iterative algorithms are used for learning. The weights are constrained to be sparse, which is required for analyzing the retrieval algorithm.

The input (query) vectors $\mathbf{x}$ are obtained from the vectors $\mathbf{y}$ of the base by additive noise: $\mathbf{x} = \mathbf{y} + \mathbf{e}$, where $\mathbf{e}$ are random sparse vectors with (bipolar) integer components. During retrieval, activity propagates first from the data neurons to the constraint neurons and then in the opposite direction, and so on for multistep retrieval. Non-linear transformations are used in neurons. In a stable state, the data neurons represent a base vector, and all constraint neurons obtain a total weighted zero input from the associated data neurons.

In [78] the vectors are divided into intersecting parts. Any part of the vector belongs to a subspace of smaller dimension than the vector dimension of that part. A subset of the constraint neurons corresponds to each part. They are not looking for an orthogonal basis of constraints, but for vectors orthogonal to the corresponding parts of the data vectors from the base: $\mathbf{W}^{(k)}\mathbf{y}^{(k)} = 0$, where $k$ is the part number. To do this, the objective function is formulated and optimized with a stochastic gradient descent (several times for each part). During retrieval, they first independently correct errors in each part by performing several steps of the network dynamics. The correction is based on the fact that

$$\mathbf{W}^{(k)}\mathbf{y}^{(k)} = \mathbf{W}^{(k)}\mathbf{e}^{(k)}.$$

Then, exploiting intersection of the parts, the parts without errors are used to correct the parts with errors.

In [145], $\mathbf{y}$ from a subspace of dimension $d < D$ are considered. Training forms a matrix $\mathbf{W}$ of $D - d$ non-zero linearly independent vectors orthogonal to the vectors $\mathbf{y}$ of the base: $\mathbf{Wy} = 0$ for all $\mathbf{y}$ of the base. An iterative algorithm of activity propagation in the network retrieves $\mathbf{y}$.

Algorithms [78, 145], described above, are claimed to store the number $N$ of vectors (generated from their respective data models) exponential in $D$ ($O(a^D)$, $a > 1$) with the possibility of correcting a number of random errors that is linear in the vector dimension, $D$. However, to ensure a high probability of retrieval, a graph with a certain structure must be obtained, which is not guaranteed by the learning algorithms used.

*NAMs based on sparse recovery algorithms.* To create autoassociative memory on the basis of a bipartite graph, in [110, 111] they use connection matrices $\mathbf{W}$ which allow them to reconstruct a sparse noise vector $\mathbf{e}$ which additively distorts the vector $\mathbf{y}$ of the base to form the query vector $\mathbf{x}$. Then the required base vector is obtained as $\mathbf{y} = \mathbf{x} - \mathbf{e}$. The noise vector is calculated using sparse recovery methods (that is, methods that find the solution vector with the least number of non-zero components). These methods require knowledge of the linear constraints matrix $\mathbf{W}$ such that $\mathbf{Wy} = 0$ for all vectors $\mathbf{y}$ of the base. For some models of vectors (i.e. constraints or generative processes for the base vectors), such $\mathbf{W}$ can be obtained in polynomial time from the base of vectors generated by the model. In contrast to [78, 145], finding $\mathbf{W}$ is guaranteed with high probability, and adversarial rather than random errors are used as noise.

In [110] real-valued vectors are used as the base, satisfying a set of non-sparse linear constraints. The data model, where the vectors of the base are given by linear combinations of vectors with sub-Gaussian components, allows storing the number of vectors $N$ up to $\exp(D^{3/4})$. The data model with a basis of orthonormal vectors provides $N$ up to $\exp(d)$, where $1 \le d \le D$. Both models allow for accurate recovery from vectors with significant noise.

In [111], as in [145], the vectors of the base are from a subspace defined by sparse linear constraints. They consider both real-valued vectors and binary vectors from $\{-1,+1\}^D$ satisfying **W** models of a certain type (sparse-sub-Gaussian model). Learning is based on solving the dictionary learning problem with a square dictionary [111]. An iterative retrieval algorithm uses the fact that **W** is an expander graph with good properties [111]. The memory capacity and resistance to distortion is increased relative to [110].

Note the drawback of the methods considered in this section is that bases of real data may not correspond to the data models used.

## DISCUSSION

In addition to being an interesting model of biological memory, neural network autoassociative distributed memories (NAMs) have also been considered as index structures that give promise to speed up nearest neighbor search relative to linear search (and, hopefully, to some other index structures). This mainly concerns sparse binary vectors of high dimension, because the number of such vectors that it is possible to memorize and retrieve from a significantly distorted version may far exceed the dimensionality of the vectors in some matrix-type NAMs, and the ratio $N/D$ may be similar to the speed-up relative to linear search (see the first four Sections).

Distributed NAMs have some drawbacks relative to traditional computer science methods for nearest neighbor search. The vector retrieved by a NAM may not be the nearest neighbor of the query vector. This could be tolerable if the output vector is an approximate nearest neighbor from the set of stored vectors. However, in NAMs the output vector may not even be a vector of the base set ("spurious memories"). (Up to a certain number of stored vectors and query vector distortion these problems remain insignificant.) For dense binary vectors, the number of vectors able to be reliably stored and retrieved is (much) smaller than the vector dimension. Also, NAMs are usually analyzed for the average case of random vectors and distortions, whereas real data are not like that, which results in poorer performance. However, available comparisons with the inverted index for sparse binary vectors in the average case do not clearly show the advantage of one or other algorithm in query time (Subsections "Hopfield networks with sparse vectors", "Willshaw networks with sparse vectors")

An obvious approach to improve the memory and time complexity of the matrix-type NAMs from quadratic to linear in vector dimension is the use of incompletely connected networks with constant (but rather large) number of connections per neuron.

An interesting direction is index structures for similarity search in which NAM modules are used at some stages. The index structure of Subsection "Willshaw networks in the index structures for nearest neighbor search" uses several NAMs to memorize parts of the base, and the similarity of the result of single-step retrieval with the query vector is used to select the "best" NAM on which to perform an exact linear search against its stored vectors.

Some studies are aimed at more efficient use of NAMs when working with real data. For example, the GB network with binary connections uses different neurons to represent the same component of the source vector, which allows for more balanced use of connections.

In Section "Generalization in NAMs" we discussed the use of NAMs for generalization rather than exact retrieval from associative memory. In the NAMs that use versions of the Hebb learning rule, storage of vectors (even random ones) is accompanied by emergence of additional stable states. For correlated vectors, their common 1-components become "tightly" connected and stable states corresponding to them arise. Revealing these stable states can be used for data mining, e.g. for binary factor analysis [36, 37]. Research of complex (possibly hierarchical) structure of stable states (discussed in terms of cores and fringes of neural assemblies) may appear useful both for modeling brain function and for applications.

Real data in many cases are not binary sparse vectors of high dimension with which the NAMs considered in the first four Sections work best. So, similarity preserving transformations to that format are required (Subsection "Similarity preserving binary vectors"). However, the obtained vectors (as well as the initial real data) are not random and independent, so the analytical and experimental results available for random, independent vectors usually can not predict NAM characteristics for real data.

Using data vectors (often non-binary) that satisfy some linear constraints (instead of random independent vectors) allows bipartite graph based NAM construction with capacity near exponential in the vector dimension (Section "NAMs with a Bipartite Graph Structure for Nonbinary Data with Constraints"). However, again, this requires data from specific vector models (to which real data often do not fit).

In NAMs with higher-order connections, connections are not between a pair, but between a larger number of neurons (this number being the order). So, the NAM becomes of tensor-type instead of matrix-type. These NAMs (Section "NAMs with Higher-Order Connections and without Connections") allow storing the number of dense vectors exponential in the order. However, this is achieved by the corresponding increase in the number of connections, and therefore in memory and in query time.

The higher-order NAMs are generalized in [95, 96], where, roughly, the sum of polynomial functions of the dot products between all memory vectors and the network state is used as the input sum of a neuron. Such a treatment makes it possible to draw interesting analogies with perceptrons and kernel methods in the classification problem. However, for nearest neighbor search, this seems impose a query time exceeding that of linear search.

Overcoming these and other drawbacks and knowledge gaps, and improving NAMs are promising topics for further research.

Let's note other directions of research in fast similarity search of binary (non- sparse) vectors. Examples of index structures for exact search are [28] (with a fixed query radius and analysis for worst-case data; however impractical due to the small query radius required for sub-linear query time and moderate memory costs) and [116] (practical, with variable radius of the query and analysis for random data).

Theoretical algorithms for approximate search (providing: sublinear search time, a specified maximum difference of the result from the result of the exact search, and no false negatives) in [2] are modifications of more practical algorithm classes related to Locality Sensitive Hashing and Locality Sensitive Filtering (see [18, 14, 147]). However, the latter allow false negatives (with low probability). Unlike NAMs, these algorithms provide guarantees for the worst-case data, but require a separate index structure for each degree of distortion of the query vector. The bounds on the ratio of binomial coefficients [2] is useful for NAMs.

We note that the index structures for the Hamming distance [28, 116] work with vectors of moderate dimension (up to hundreds), and for binary sparse high dimensional vectors Jaccard similarity index structures are used [147, 2, 27, 29]. A survey of these and other similar index structures is presented in the forthcoming [133], see also [132] for another type of index structures.

*В.И. Гриценко[1]*, член-корреспондент НАН Украины директор,
e-mail: vig@irtc.org.ua
*Д.А. Рачковский[1]*, д-р техн. наук, вед. науч. сотр.
отд. нейросетевых технологий обработки информации,
e-mail: dar@infrm.kiev.ua
*А.А. Фролов[2]*, д-р биол.наук, проф.,
факультет электротехники и информатики,
e-mail: docfact@gmail.com
*Р. Гейлер[3]* PhD (психология), исследователь,
e-mail: r.gayler@gmail.com
*Д. Клейко[4]*, аспирант, факультет информатики,
электрической и космической техники.
e-mail: denis.kleyko@ltu.se
*Е. Осипов[4]*, PhD (информатика), проф., факультет информатики,
электрической и космической техники
e-mail: evgeny.osipov@ltu.se
[1] Международный научно-учебный центр информационных технологий
и систем НАН Украины и МОН Украины, пр. Академика Глушкова, 40,
г. Киев, 03187, Украина
[2] Технический университет Остравы, 17 listopadu 15,
708 33 Острава-Поруба, Чешская Республика
[3] Мельбурн, штат Виктория, Австралия
[4] Технологический университет Лулео, 971 87 Лулео, Швеция

## НЕЙРОСЕТЕВАЯ РАСПРЕДЕЛЕННАЯ АВТОАССОЦИАТИВНАЯ ПАМЯТЬ: ОБЗОР

В настоящем обзоре рассмотрены модели автоассоциативной распределенной памяти, которые могут быть естественным образом реализованы нейронными сетями. Модели используют для запоминания векторов в основном локальном правиле обучения путем модификации значений весов межнейронных связей, которые существуют между всеми нейронами (полносвязные сети). В распределенной памяти различные векторы запоминают в одних и тех же ячейках памяти, которым в рассматриваемом случае нейронной сети соответствуют одни и те же связи. Обычно исследуют запоминание векторов, случайно выбранных из некоторого распределения.

При подаче на вход автоассоциативной памяти искаженных вариантов запомненных в ней векторов осуществляется извлечение (восстановление) ближайшего запомненного вектора. Это реализуется за счет итеративной динамики нейронной сети на основе локально доступной в нейронах информации, полученной по связям от других нейронов сети. Вплоть до определенного количества запомненных в сети векторов и степени их искажения на входе, в результате динамики сеть с симметричными связями приходит в устойчивое состояние, соответствующее запомненному в сети вектору, имеющему наибольшее сходство с входным вектором (сходство обычно измеряют в терминах скалярного произведения).

Такие нейросетевые варианты автоассоциативной памяти позволяют запомнить с возможностью восстановления такого количества векторов, которое может превышать размерность векторов (совпадающую с количества нейронов в сети). Для векторов большой размерности это открывает возможность поиска приближенного ближайшего соседа с временной сложностью, сублинейной от количества запомненных в нейронной сети векторов. К недостаткам такой памяти относится то, что восстановленный динамикой сети вектор может не быть ближайшим ко входному или даже может вообще не принадлежать к множеству запомненных векторов и значительно отличаться от любого из них. Исследования различных типов нейросетевой автоассоциативной памяти направлены на выявление диапазонов параметров, при которых указанные недостатки проявляются с малой вероятностью, а достоинства выражены в максимальной степени.

*V.I. Gritsenko, D.A. Rachkovskij, A.A. Frolov, R. Gayler, D. Kleyko, E. Osipov*

Основное внимание уделено сетям с парными связями типа Hopfield, Willshaw, Potts и работе с бинарными разреженными векторами (векторами с количеством единичных компонентов, малым по сравнению с количеством их нулевых компонентов), т.к. только для таких векторов удается запомнить с возможностью восстановления большое количество векторов. Помимо функции автоассоциативной памяти, для этих сетей также обсуждается функция обобщения. Обсуждаются также неполносвязные сети. Кроме того, рассмотрена автоассоциативная память в нейронных сетях со связями высшего порядка — то есть со связями не между парами, а между большим количеством нейронов.

Рассмотрена также автоассоциативная память в нейронных сетях со структурой двудольного графа, где одно множество нейронов представляет запоминаемые векторы, а другое — линейные ограничения, которым они подчиняются. Эти сети выполняют функцию автоассоциативной памяти и для небинарных данных, удовлетворяющих заданной модели ограничений.

Обсуждаются отношение рассмотренных в обзоре моделей нейросетевой автоассоциативной распределенной памяти к проблематике поиска по сходству, достоинства и недостатки рассмотренных методов, направления дальнейших исследований. Один из интересных и все еще не полностью разрешенных вопросов заключается в том, может ли нейронная автоассоциативная память искать приближенных ближайших соседей быстрее других индексных структур для поиска по сходству, в частности, для случая векторов очень больших размерностей.

*Ключевые слова:* *распределенная ассоциативная память, разреженный бинарный вектор, сеть Хопфилда, память Уиллшоу, модель Поттса, ближайший сосед, поиск по сходству.*

*В.І. Гриценко[1]*, член-кореспондент НАН України, директор,
e-mail: vig@irtc.org.ua
*Д.А. Рачковський[1]*, д-р техн. наук, пров. наук. співроб.
відд. нейромережевих технологій оброблення інформації,
e-mail: dar@infrm.kiev.ua
*А.А. Фролов[2]*, д-р біол. наук, проф.,
факультет електротехніки та інформатики,
e-mail: docfact@gmail.com
*Р. Гейлер[3]*, PhD (психологія), дослідник,
e-mail: r.gayler@gmail.com
*Д. Клейко[4]*, аспірант, факультет інформатики,
електричної та космічної техніки.
e-mail: denis.kleyko@ltu.se
*Е. Осипов[4]*, PhD (інформатика), проф., факультет інформатики,
електричної та космічної техніки
e-mail: evgeny.osipov@ltu.se

[1] Міжнародний науково-учбовий центр інформаціоних технологій
та систем НАН України та МОН України, пр. Академіка Глушкова, 40,
м. Київ, 03187, Україна
[2] Технічний університет Острави, 17 listopadu 15,
708 33 Острава-Поруба, Чеська Республіка
[3] Мельбурн, штат Вікторія, Австралія
[4] Технологічний університет Лулео, 971 87 Лулео, Швеція

## НЕЙРОМЕРЕЖНА РОЗПОДІЛЕНА АВТОАССОЦІАТИВНА ПАМ'ЯТЬ: ОГЛЯД

У цьому огляді розглянуто моделі автоасоціативної розподіленої пам'яті, які можуть бути природним чином реалізовані нейронними мережами. Моделі використовують для запам'ятовування векторів в основному локальному правилі навчання шляхом модифікації значень ваг міжнейронних зв'язків, які існують між всіма нейронами (повнозв'язні мережі). У розподіленій пам'яті різні вектори запам'ятовуються в одних і тих самих елементах пам'яті, яким в цьому випадку нейронної мережі відповідають одні і ті ж зв'язки. Зазвичай досліджують запам'ятовування векторів, випадково вибраних з деякого розподілу.

Якщо на вхід автоасоціативної пам'яті подаються спотворені варіанти запам'ятованих в ній векторів, здійснюється витяг (відновлення) найближчого раніше запам'ятованого вектора. Це реалізується за рахунок ітераційної динаміки нейронної мережі на основі локально доступної в нейронах інформації, отриманої від інших нейронів мережі. До певної кількості запам'ятованих в мережі векторів і ступеня їх спотворення на вході, в результаті динаміки мережа із симетричними зв'язками приходить в стійкий стан, відповідний запам'ятованому в мережі вектору, який має найбільшу схожість з вхідним вектором (схожість зазвичай вимірюють як скалярний добуток).

Такі нейромережні варіанти автоасоціативної пам'яті дозволяють запам'ятати з можливістю відновлення таку кількість векторів, яка може перевищувати розмірність векторів (що збігається з кількістю нейронів в мережі). Для векторів великої розмірності це відкриває можливість пошуку наближеного найближчого сусіда з складністю, сублінійною від кількості запам'ятованих в нейронній мережі векторів. До недоліків такої пам'яті відноситься те, що відновлений динамікою мережі вектор може не бути найближчим до вхідного або навіть може взагалі не належати до множини запам'ятованих векторів і значно відрізнятися від будь-якого з них. Дослідження різних типів нейромережної автоасоціативної пам'яті спрямовано на виявлення діапазонів параметрів, при яких зазначені недоліки проявляються з малою імовірністю, а достоїнства виражені в максимальному ступені.

Основну увагу приділено мережам з парними зв'язками типу Hopfield, Willshaw, Potts і роботі з бінарними розрідженими векторами (векторами з кількістю одиничних компонентів, яке є малим у порівнянні з кількістю їх нульових компонентів), так як тільки для таких векторів вдається запам'ятати з можливістю відновлення велику кількість векторів. Крім функції автоасоціативної пам'яті, для цих мереж також обговорюється функція узагальнення. Обговорюються також неповнозв'язкові мережі. Крім того, розглянуто автоасоціативну пам'ять в нейронних мережах зі зв'язками вищого порядку — тобто зі зв'язками не між парами, а між великою кількістю нейронів.

Розглянуто також автоасоціативна пам'ять в нейронних мережах зі структурою двудольного графа, де одна множина нейронів надає вектори, які запам'ятовуються, а інша — лінійні обмеження, яким вони підкоряються. Ці мережі виконують функцію автоасоціативної пам'яті також для небінарних даних, які відповідають заданій моделі обмежень.

Обговорюються можливості використання розглянутих в огляді моделей нейромережної автоасоціативної розподіленої пам'яті у проблематиці пошуку за схожістю, достоїнства і недоліки розглянутих методів, напрямки подальших досліджень. Один із цікавих і все ще не повністю вирішених питань полягає в тому, чи може нейронна автоасоціативна пам'ять шукати наближених найближчих сусідів швидше інших індексних структур для пошуку за схожістю, зокрема, у випадку векторів дуже великих розмірностей.

*Ключові слова: розподілена асоціативна пам'ять, розріджений бінарний вектор, мережа Хопфілда, пам'ять Уілшоу, модель Потса, найближчий сусід, пошук за схожістю.*